\documentclass[twoside]{article}

\usepackage[accepted]{aistats2022}
\usepackage{amsmath,graphicx}

\usepackage{url,amsfonts,amssymb,bm}

\usepackage{enumitem}
\usepackage{algorithm}
\usepackage{algorithmicx, algpseudocode }
\usepackage{multirow}
\usepackage{color}
\usepackage{mathtools}

\usepackage{caption}
\usepackage{subcaption}
\usepackage{amsthm}

\usepackage[colorlinks=true, linkcolor=magenta, citecolor=blue, urlcolor=blue]{hyperref}
\hypersetup{pdfcreator  = {LaTeX with hyperref package},
               pdfproducer = {dvips + ps2pdf} }

\newtheorem{remark}{Remark}

\newtheorem{theorem}{Theorem}
\newtheorem{lemma}{Lemma}

\newtheorem{proposition}{Proposition}
\newtheorem{assumption}{Assumption}
\usepackage{bm}

\long\def\comment#1{}

\newfont{\bbb}{msbm10 scaled 700}

\newcommand{\E}{{\mathbb E}}
\newcommand{\R}{{\mathbb R}}

\newcommand{\bv}{{\bf b}}

\newcommand{\gv}{{\bf g}}

\newcommand{\uv}{{\bf u}}
\newcommand{\wv}{{\bf w}}

\newcommand{\xv}{{\bf x}}

\newcommand{\Am}{{\bf A}}

\newcommand{\Cm}{{\bf C}}

\newcommand{\Id}{{\bf I}}
\newcommand{\Jm}{{\bf J}}
\newcommand{\Km}{{\bf K}}
\newcommand{\Lm}{{\bf L}}

\newcommand{\Sm}{{\bf S}}

\newcommand{\Wm}{{\bf W}}

\newcommand{\Xm}{{\bf X}}

\newcommand{\Ec}{{\cal E}}

\newcommand{\Gc}{{\cal G}}

\newcommand{\Lc}{{\cal L}}

\newcommand{\Oc}{{\cal O}}

\newcommand{\Sc}{{\cal S}}
\newcommand{\Tc}{{\cal T}}

\newcommand{\Vc}{{\cal V}}

\newcommand{\Lambdam}{\hbox{\boldmath$\Lambda$}}

\newcommand{\Sigmam}{\hbox{\boldmath$\Sigma$}}

\newcommand{\Thetam}{\hbox{\boldmath$\Theta$}}

\newcommand{\diag}{{\hbox{diag}}}
\renewcommand{\det}{{\hbox{det}}}

 \DeclareMathOperator*{\argmin}{arg\,min}
  \DeclareMathOperator*{\rank}{{rank}}

  \DeclareMathOperator*{\tr}{tr}

\DeclareMathOperator*{\Prob}{\mathbb{P}}
\usepackage[round]{natbib}

\begin{document}



\twocolumn[

\aistatstitle{Laplacian Constrained Precision Matrix Estimation: Existence and  High Dimensional Consistency}

\aistatsauthor{ Eduardo Pavez  }

\aistatsaddress{  University of Southern California \\ Los Angeles, CA } ]

\begin{abstract}
  This paper considers the problem of estimating high dimensional Laplacian constrained precision matrices by minimizing  Stein's loss. We obtain a necessary and sufficient  condition for existence of this  estimator, that  consists on checking whether a certain data dependent graph is connected. We also prove consistency in the high dimensional setting under the symmetrized Stein loss. We show that the error rate does not depend on the graph sparsity, or other type of structure, and that Laplacian constraints are sufficient for high dimensional consistency.  Our proofs  exploit properties of graph Laplacians, the matrix tree theorem, and a characterization of the proposed estimator based on effective graph resistances. We validate our theoretical claims with numerical experiments.
\end{abstract}

\section{Introduction}
\label{sec:intro}

Graph based algorithms are becoming increasingly popular in application domains where unstructured and irregular  data are pervasive,  including signal and image processing \citep{shuman2013emerging,ortega2018graph},  and machine learning \citep{von2007tutorial,bronstein2017geometric}.  

In graph based data analysis,  observations are located at the  nodes (or vertices), while the graph structure and edge weights represent affinity relations between the nodes. 
A fundamental problem  is  estimation of graph structure and edge weights from nodal observations.   Earlier   approaches were based on sparse inverse covariance matrix estimation \citep{dempster1972covariance,lauritzen1996graphical,friedman2008sparse,rothman2008sparse}. More recently, there has been growing interest in estimating  Laplacian constrained inverse covariance (precision) matrices \citep{lake2010discovering,slawski2015estimation,egilmez2017graph,dong2019learning,mateos2019connecting,tugnait2021sparse}. This approach can benefit from developed statistical and graph theories. For example,  in the Gaussian case, the entries of the precision matrix correspond to partial correlations \citep{lauritzen1996graphical}, while additionally,  the spectral decomposition of the Laplacian can be interpreted as a Fourier basis for signals on graphs \citep{shuman2013emerging,ortega2018graph,ortega2021book}.

In this work, we consider undirected graphs with $p$ nodes, and study  estimation of the combinatorial graph  Laplacian (CGL) matrix  from signals $\lbrace \xv_1,\cdots, \xv_n\rbrace \subset \R^p$, via the  CGL estimator (CGLE):
\begin{equation}\label{eq_optimal_proposed}
     \widehat{\Lm}=\argmin_{\Lm \in \Lc_p(\Ec)}  -\log\det^{\dagger}\left(\Lm \right) + \tr(\Lm \Sm).
\end{equation}
 Here $\Lc_p(\Ec)$ is the set of  CGL matrices of connected graphs with edge set contained in $\Ec$, the product of non zero eigenvalues is denoted by $\det^{\dagger}$, and  $\Sm \coloneqq  n^{-1}\sum_{i=1}^n\xv_i \xv_i^{\top}$ is the sample covariance matrix.
%defined as
% \begin{equation}
% \Sm \coloneqq  n^{-1}\sum_{i=1}^n\xv_i \xv_i^{\top}.
% \end{equation}
When the signals $\lbrace  \xv_i \rbrace$ are i.i.d., with zero mean and population covariance matrix $\Sigmam^{\star} =(\Lm^{\star})^{\dagger} $,  \eqref{eq_optimal_proposed} is the maximum likelihood estimator   of a  Gaussian Markov random field   with Laplacian constrained precision matrix (L-GMRF)  \citep{lauritzen1996graphical,egilmez2017graph,ying2020nonconvex}.  This estimator also promotes smooth signal representations, essential for signal processing and learning on graphs \citep{dong2020graph,hu2021graph}.%, while also having useful geometric properties and  efficient estimation algorithms \citep{pavez2019efficient,zhao2019optimization,kumar2020unified}.

In applications such as portfolio selection and gene expression analysis \citep{cai2011constrained,agrawal2019covariance}, the dimension $p$ can be much larger than the sample size $n$. Hence, estimators that perform well and are consistent in this regime   have become increasingly attractive. Traditional precision matrix estimation exploited sparsity to achieve high dimensional consistency, and guarantee existence of the precision matrix estimator \citep{cai2011constrained,rothman2008sparse}. It has been shown that when Laplacian constraints are used, various Laplacian estimators exist under milder conditions  \citep{slawski2015estimation, ying2021minimax}. Consistency in the high dimensional setting has been proven for the generalized graph Laplacian estimator \citep{soloff2020covariance}, without sparsity assumptions. While for the CGLE case, available results assume graph sparsity \citep{rabbat2017inferring,ying2020nonconvex,ying2021minimax}.

 In this paper we study existence and high dimensional consistency of the CGLE. Our contributions can be summarized as follows:
\begin{itemize}
\itemsep0em 
    \item We obtain  a necessary and sufficient condition for existence of the CGLE (\autoref{th_existence}). This condition reduces to checking  whether  a certain data dependent graph is connected. %The necessary condition obtained by \cite{ying2021minimax} is a special case of .
     \item We  prove  that the CGLE is  consistent in the high dimensional setting ($p>n$) under the symmetrized Stein loss \citep{stein1975estimation,ledoit2018optimal} with a rate $\Oc(\sqrt{\log(p)/n})$ (\autoref{th_consistency}).   %
     \item We prove that  tree structured graphs can be estimated at the faster rate  $\Oc(\log(p)/n)$ (\autoref{th_consistency_trees}).
\end{itemize}
  Our proofs are relatively simpler than recent works studying existence of the CGLE, and  consistency   in the Frobenius norm \citep{ying2020nonconvex,ying2021minimax}. An essential part of our analysis is based on  a characterization of the CGLE by \citet{pavez2019efficient}  that uses  effective graph  resistances and the matrix tree theorem \citep{klein1993resistance,xiao2003resistance,ellens2011effective}.  
  Unlike other Laplacian and   (inverse) covariance estimators that require some structural assumption such as sparsity, bounded degree, or bandable structures to achieve high dimensional consistency \citep{cai2016estimating,rabbat2017inferring,ying2020nonconvex,ying2021minimax},  our results are able to achieve the rate $\Oc(\sqrt{\log(p)/n})$   by exploiting properties of CGL matrices and using the symmetrized Stein loss,   without requiring other types of graph structure.  
  
  The rest of the paper is organized as follows. In Section \ref{sec:related} we review related work. We  introduce notation, and some graph theoretic concepts in Section \ref{sec:prel}. Our results about existence and consistency are presented in Sections \ref{sec:exist}  and \ref{sec:consis}, respectively. We finalize with numerical experiments in Section \ref{sec:exp}, and conclusions in Section \ref{sec:conc}. 
%This and related  graph Laplacian estimators  have become ubiquitous in  graph signal processing applications   \citep{ortega2018graph,egilmez2017graph,egilmez2018graph}.
%%%%     Related estimators
\begin{table*}[htb]
\caption{Comparison of key parameters for estimation of Laplacian constrained precision matrices. GGL are the matrices $\lbrace \Thetam \succ 0$, $\theta_{ij} \leq 0,~  \forall i \neq j\rbrace$. The sample correlation and distant coefficients are  $\hat{\rho}_{ij} = S_{ij}/\sqrt{S_{ii}S_{jj}}$, and  $h_{ij} =S_{ii} + S_{jj} - 2S_{ij} $ respectively.  $\rho$ and $\hat{\rho}$ are the largest  population and sample correlation coefficients respectively, thus $\rho = \max_{i\neq j} \Sigma^{\star}_{ij}/\sqrt{\Sigma^{\star}_{ii} \Sigma^{\star}_{jj}}$, and $\hat{\rho} = \max_{i \neq j} \hat{\rho}_{ij}$. }
\label{tab_existence}
\begin{center}
\scalebox{0.9}{
\begin{tabular}{| l l l l|}
\hline
	Estimator    & Existence condition 		&   Optimal weights	& Convergence rate \\ \hline 
	GGL & $\hat{\rho} < 1 $ \citep{slawski2015estimation}& $|\hat{\theta}_{ij}| \leq \frac{|S_{ij}|/S_{ii}S_{jj}}{1 - \hat{\rho}^2_{ij}}$ (\autoref{ap_bound_GGL})  & $\frac{1}{1-\rho} \sqrt{\frac{\log(p)}{n}}$ \citep{soloff2020covariance} \\ %\hline
	CGL   & $\min_{(i,j) \in \Ec} h_{ij}>0$ (Th. \ref{th_existence})& $\hat{w}_{ij} \leq \frac{1}{h_{ij}} $ \citep{pavez2019efficient} & $ \sqrt{\frac{\log(p)}{n}}$ (Th. \ref{th_consistency}) \\ \hline 
\end{tabular}
}
\end{center}
\end{table*}
\section{Related work}
\label{sec:related}
\paragraph{Inverse covariance estimators.} 
The CGLE   is closely related to  sparse inverse covariance matrix estimators of the form:
\begin{equation}\label{eq_ML_regularized_precision}
    \min_{\Thetam \in \Sc}-\log\det(\Thetam) + \tr(\Thetam \Sm) + \alpha \Vert \Thetam \Vert_1.
\end{equation}
When $\Sc$ is the set of symmetric positive definite matrices and $\alpha=0$,  \eqref{eq_ML_regularized_precision} is the maximum likelihood  estimator (MLE)  of the inverse covariance matrix under a  Gaussian distribution  \citep{dempster1972covariance}. 
When $\alpha>0$, this estimator is known as   graphical lasso \citep{friedman2008sparse}. 
It was recently   proposed to solve \eqref{eq_ML_regularized_precision}  for matrices with non positive off diagonal entries, i.e., $\Sc$ is the set of M-matrices, also known as generalized graph Laplacians (GGL). The unregularized case ($\alpha=0$ ) was first studied by   \citet{slawski2015estimation} and the $\ell_1$ regularized ($\alpha>0$) by \citet{egilmez2017graph,pavez2018learning}. The estimator \eqref{eq_optimal_proposed} was first proposed by \citet{egilmez2017graph}. Since then, several algorithms and extensions  for estimating CGL matrices have been proposed \citep{hassan2016topology,hassanmoghaddam2017topology,rabbat2017inferring,egilmez2018graph,pavez2019efficient,zhao2019optimization,kumar2020unified}.
%
%
%\noindent
\paragraph{Existence.}  
The Gaussian MLE does not exist when $n<p$, unless additional sparsity constraints are imposed on $\Thetam$, in which case, existence is guaranteed for $\tau \leq n <p$, for some $\tau>0$ that depends on the graph structure \citep{uhler2012geometry}. The graphical lasso always exists for every $n\geq 1$,  however note that using this estimator is only reasonable when sparsity is required. 
In contrast, the generalized graph Laplacian  estimator (GGLE) and the CGLE  exist with probability one, as long as $n \geq 2$ under mild conditions on the data distribution \citep{slawski2015estimation,ying2021minimax}. In particular, the CGLE exists with high probability when  $\Ec$ is the complete graph and the data follows a Laplacian Gaussian Markov random field  distribution \citep{ying2021minimax}. Unlike previous work,  we obtain  a necessary and sufficient  condition for existence of the CGLE for the more general case that includes edge constraints.
\paragraph{Consistency and  regularization.}
It is  well known that for many optimization problems, adding  $\ell_1$ regularization produces sparse solutions. In particular, as the regularization parameter $\alpha$  is increased,    the regularized MLE solution \eqref{eq_ML_regularized_precision} becomes more sparse, for  both the graphical Lasso \citep{witten2011new,mazumder2012exact,hsieh2012divide} and  GGL \citep{pavez2018learning,lauritzen2019maximum} estimators. The opposite is observed when    an $\ell_1$ norm term   with large regularization parameter is added to the CGLE , in which case   the graph becomes more dense \citep{ying2020nonconvex}. Various non convex  alternatives to the $\ell_1$ norm  have been proposed to overcome this issue and obtain sparse graphs \citep{ying2020nonconvex,koyakumaru2021graph,zhang2020learning,ying2021minimax}.
Besides aiding with existence,  sparse regularization can help proving consistency of  (inverse) covariance estimators in high dimensions \citep{rothman2008sparse,cai2011constrained}. In contrast,  sparsity or other types of structure, are not necessary for  consistency, even in high dimensions when GGL  \citep{soloff2020covariance} or CGL (\autoref{th_consistency}) constraints are used.   
\paragraph{Comparison with the GGL estimator.} 
 Given that  the CGL matrix is just a GGL whose rows sum to zero (i.e.,  $\Lm \cdot  \mathbf{1=0}$), the reader may ask  if there are meaningful differences between the GGL and CGL estimators. From a graph theoretic perspective, the CGL is the most widely studied graph matrix, and the relation between its spectral and vertex domain properties are better understood. On the other hand, GGL matrices provide a richer parameter space \citep{egilmez2017graph}. The  GGL and CGL estimators also  have very different properties. In \autoref{tab_existence} we list some of these differences. For instance, the behaviour of the GGLE is completely characterized by correlation coefficients between pairs of nodes, while the CGLE is characterized by node distances.  \citet{lu2018learning} gives  examples illustrating differences between GGLE and CGLE on tree structured graphs. Another major difference between both estimators is their interaction with   $\ell_1$ regularization. While the GGLE becomes sparse as the regularization increases \citep{pavez2018learning}, the CGLE becomes more dense \citep{ying2020nonconvex}.
%
%
%
%\begin{table}[ht]
%\centering
%\scalebox{1}{
%\begin{tabular}{|l|l|l|}
%\hline
%	Ref.    & Existence 		&   Assumption	 \\ \hline 
%	MLE \citep{uhler2012geometry} & $n > \tau$   & Graph structure with $\tau<p$   \\ \hline
%	Glasso \citep{friedman2008sparse} & $n \geq 1$ & $\alpha>0$  \\ \hline 
%	GGLE \citep{slawski2015estimation} & $n \geq 2$ & $\max_{i\neq j}\frac{S_{ij}}{\sqrt{S_{ii}S_{jj}}} < 1 $, and $\min_i S_{ii}> 0$ \\ \hline
%	CGLE \citep{ying2021minimax} & $n \geq 1$ &   L-GMRF distribution \\ \hline \hline 
% 	CGLE  Th. \ref{th_existence}, \ref{th_existence_converse}& $n \geq 1$ &  $\Ec, \Ec_{\exists}$ are connected, and $\Ec \subset \Ec_{\exists}$\\ \hline 
%\end{tabular}
%}
%\caption{Existence conditions for inverse covariance matrix estimators.}
%\label{tab_existence}
%\end{table}
%
%
\section{Preliminaries}
\label{sec:prel}
%\subsection{Notation}
We denote scalars, vectors and matrices using lower case, lower case bold, and upper case bold fonts, respectively, e.g.., $a, \bv, \Cm$. For vectors $\Vert \bv\Vert_q$ denotes the $\ell_q$ norm.
The set of integers $\lbrace 1,2,\cdots, q\rbrace$ is often denoted by $[q]$. Variables denoted with hat, e.g., $\hat{a}, \hat{\bv}, \widehat{\Cm}$ correspond to estimated quantities, while the use of $\star$, indicates a population quantity, e.g.,   ${a}^{\star}, {\bv}^{\star}, {\Cm}^{\star}$. For matrices, $\Am^{-1}$ indicates inverse, while $\Am^{\dagger}$ is the Moore-Penrose pseudo inverse. 

%\subsection{Graphs}
A graph  $\Gc=(\Vc,\Ec, \mathbf{W})$ is a triplet consisting of a node set $\mathcal{V} = \lbrace 1,\cdots, p \rbrace$, an  edge set $\mathcal{E} \subset \mathcal{V} \times \mathcal{V}$, and edge weights stored in the matrix $\Wm$.  Individual edges will be denoted by $e = (i,j) \in \mathcal{E}$, and their corresponding edge weight by $w_e $ or $ w_{ij}$ depending on the context. We consider undirected graphs with  positive weights, so the weight matrix $\mathbf{W} = (w_{ij})$ is symmetric and  non negative, and  $w_{ij} = w_{ji} > 0$ if and only if $e = (i,j) \in \mathcal{E}$, in this case  $(j,i)$ is not on the edge set. 
We also denote by $\wv = [w_{e_1}, \cdots, w_{e_m}]^{\top}$ the $m$ dimensional vector containing the edge weights, where $e_i \in \Ec$. The degree of a node $i$ is defined as $d_i = \sum_{j} w_{ij}$, and the degree matrix is $\mathbf{D} = \diag([d_1, d_2, \cdots, d_p])$. 
 We denote  the edge vector $\mathbf{g}_e \in \mathbb{R}^p$, which has entries $\mathbf{g}_e(i) =1$,  $\mathbf{g}_e(j) = -1$, and zero otherwise. The incidence matrix is defined as $\mathbf{\Xi} = [\mathbf{g}_{e_1}, \mathbf{g}_{e_2}, \cdots, \mathbf{g}_{e_m}]$, and the CGL matrix of $\Gc$ is  
 \begin{equation}\label{eq_cgl_definition}
 \mathbf{L} \coloneqq \mathbf{D -W} = \mathbf{\Xi \diag( w) \Xi^{\top}} = \sum_{e \in \Ec} w_e \gv_e \gv_e^{\top}.
 \end{equation}
 The CGL is symmetric and positive semi-definite, with eigenvalues  $0=\lambda_1(\Lm) \leq \lambda_2(\Lm) \leq \cdots \leq \lambda_p(\Lm)$. For connected graphs,  $\lambda_2(\Lm)$ is always positive  \citep{fiedler1973algebraic}. %Disconnected graphs have algebraic connectivity  equal to zero, while better connected graphs have a   larger algebraic connectivity.   
 Let $\Jm_p$ be  is the all ones $p \times p$ matrix.
 The following statements are equivalent to a graph being connected: 1) $\lambda_2(\Lm)>0$, 2) $\rank(\Lm)=p-1$, 3) the matrix $\Lm + (1/p)\Jm_p$ is positive definite, and 4) $\det^{\dagger}(\Lm) = \det(\Lm + (1/p)\Jm_p) >0$.
%  \begin{enumerate}
%      %\item a graph with CGL $\Lm$ is connected,
%      \item $\lambda_2(\Lm)>0$, 
%      \item $\rank(\Lm)=p-1$,
%      \item 
%      \item $\det^{\dagger}(\Lm) = \det(\Lm + (1/p)\Jm_p) >0$
%  \end{enumerate}
 The set of CGL matrices of connected graphs with edge set contained in  $\Ec$ is denoted by
 \begin{equation}
 \Lc_{p}(\Ec) = \left\lbrace \Lm: \Lm =  \sum_{e \in \Ec} w_e \gv_e \gv_e^{\top}, \wv \geq \mathbf{0}, \lambda_2(\Lm)> 0 \right\rbrace.
 \end{equation}
 The \emph{effective resistance} of edge $e=(i,j)$ is defined as
\begin{equation}\label{eq_eff_resist_defin}
r_e = \gv_{e}^{\top} \Lm^{\dagger} \gv_e = \gv_{e}^{\top} (\Lm + ({1}/{p})\Jm_p)^{-1} \gv_e.
\end{equation}
Note that when $\Sigmam = \Lm^{\dagger}$, we also have $r_e= \Sigma_{ii} + \Sigma_{jj} - 2\Sigma_{ij}$.
%Let $\rv = [r_{e_1}, \cdots, r_{e_m}]^{\top}$ be the $m$  vector containing the effective resistances of edges in $\Ec$. 
The effective resistance forms a metric on the graph. For any pair of vertices $r_{ij}=0$ when $i=j$, and $r_{ij}>0$ for any $i \neq j$ when the graph is connected. Also, they obey two triangular inequalities, namely, for any triple of nodes $i,j,k$, $r_{ij} \leq r_{ik} + r_{kj}$, and $\sqrt{r_{ij}} \leq \sqrt{r_{ik}} + \sqrt{r_{kj}}$   \citep{klein1993resistance,xiao2003resistance,ellens2011effective}.
 \section{Existence}
\label{sec:exist}
  Because the CGLE in \eqref{eq_optimal_proposed} is defined using maximum likelihood, we can define  a variation of   Stein's loss for CGL matrices of connected graphs. Let $\Lm$ and $\Lm^{\star}$ be the CGL of  connected graphs,  Stein's loss is given by
\begin{align*}
    &\ell^s(\Lm, \Lm^{\star})\coloneqq  \\
    &\frac{\left(\tr(\Lm (\Lm^{\star})^{\dagger})  - \log\det^{\dagger}(\Lm ) + \log\det^{\dagger}(\Lm^{\star}) \right)}{p-1} - 1.
\end{align*}
 Since CGL matrices of connected graphs have rank $p-1$,   instead of normalizing by $p$ as it is done for full rank matrices \citep{stein1975estimation,ledoit2018optimal,soloff2020covariance},  we  normalize by $p-1$.  With this notation,   \eqref{eq_optimal_proposed} is equivalent to
 \begin{equation}
     \min_{\Lm} \ell^s(\Lm,\Sm^{\dagger})\textnormal{ s.t. } \Lm \in \Lc_p(\Ec).
 \end{equation}
 Stein's loss is non negative when both arguments belong to $\Lc_p(\Ec)$,  and  $\ell^s(\Lm_1,\Lm_2) = 0$ if and only if $\Lm_1 = \Lm_2$. However, since $\Sm^{\dagger} \notin \Lc_p(\Ec)$, the objective function $\ell^s(\Lm,\Sm^{\dagger})$ might be unbounded below, and   $\eqref{eq_optimal_proposed}$ may not have a solution. 
%In this section we derive   a condition under which   $\ell^s(\Lm,\Sm^{\dagger})$ is bounded below, and thus the CGLE exists.
Consider the undirected  graph
\begin{equation}\label{eq_existence}
    \Ec_{\exists} = \left\lbrace (i,j) :\frac{1}{n}\sum_{k=1}^n (\xv_k(i) -\xv_k(j))^2 > 0 \right\rbrace,
\end{equation}
where $\xv_{k}(i)$ is the $i$th entry of $\xv_k$.
\begin{theorem}\label{th_existence}
The CGLE exists if and only if $\Ec_{\exists}$ is connected and $\Ec \subset \Ec_{\exists}$.
%For any connected edge set $\Ec$, the CGLE exists if and only if $\Ec \subset \Ec_{\exists}$  and $\Ec_{\exists}$ is connected. 
\end{theorem}
Note that when the  $\lbrace \xv_i \rbrace$ follow a L-GMRF distribution, the graph $\Ec_{\exists}$ is connected with probability one as long as $n\geq 1$, thus    Theorem 1 by \citet{ying2021minimax} is a special case of  \autoref{th_existence}. 

\citet{pavez2019efficient} showed that the optimal edge weights of the CGLE satisfy $\hat{w}_e \leq 1/h_{e},\textnormal{ for all } e \in \Ec$,
where we define 
\begin{equation}\label{eq_he}
    h_e = \frac{1}{n}\sum_{k=1}^n (\xv_k(i) -\xv_k(j))^2.
\end{equation}
Essentially, the condition that $\Ec \subset \Ec_{\exists}$ and $\Ec_{\exists}$ is connected, guarantees that the edge weights are bounded.
\subsection{Proof of \autoref{th_existence}}
 The proof relies on some properties of CGL matrices. Let $\Gc=(\Vc,\Ec,\Wm)$ be a connected graph, then the weight of a spanning tree $\Tc \subset \Ec$, and the weight of a graph  $\Gc$ with CGL matrix $\Lm$, are defined as
\begin{align}
\Omega(\Lm_{\Tc}) \coloneqq \prod_{e \in \Tc}w_e,~ \textnormal{ and }~ \Omega(\Lm) \coloneqq \sum_{\Tc \subset \Ec}\Omega(\Lm_{\Tc}),
\end{align}
respectively.
%  The weight of the graph $\Gc$ with CGL matrix $\Lm$   is the sum of the weights of its spanning trees, 
% \begin{align}
    % \Omega(\Lm) \coloneqq \sum_{\Tc \subset \Ec}\Omega(\Lm_{\Tc}).
% \end{align}
For unweighted graphs (all $w_e=1$), trees have unit weight, therefore the weight of a graph is the  number of its spanning trees, denoted by $\Omega(\Ec)$.
The matrix-tree theorem relates the number of spanning trees to the determinant of $\Lm$. 
\begin{theorem}\label{th_matrix_tree} The number of spanning trees satisfies
\begin{equation}\label{eq_matrix_Tree}
   \Omega(\Lm) = (1/p) \det\left(  \mathbf{L} + ({1}/{p})\mathbf{J}_p\right).
\end{equation}
\end{theorem}
 By using effective resistances and the matrix tree Theorem, we can prove \autoref{th_existence} using simple arguments.  We divide the proof   in two parts, one for each direction of the equivalence.

\textbf{Part 1:}
First we assume that the CGLE, denoted by $\widehat{\Lm}$,  exists. This  requires  the edge set $\Ec$ to be connected. If $\Ec$ is not connected, $\Omega(\widehat{\Lm})=0$, and \eqref{eq_optimal_proposed} is unfeasible. Also, 
 $\widehat{\Lm}$ satisfies the KKT optimality conditions \citep{pavez2019efficient}, which state that for all $e \in \Ec$:
\begin{align}\label{eq_kkt_1}
    -\hat{r}_e + h_e - \hat{\lambda}_e &= 0, \\ \label{eq_kkt_2}
    \hat{w}_e \hat{\lambda}_e &=0, \\ \label{eq_kkt_3}
    \hat{w}_e,~\hat{\lambda}_e &\geq 0,
\end{align}
where $\hat{r}_e$ and $\hat{w}_e$ are the optimal effective resistance and weight of edge $e$ respectively, and $\hat{\lambda}_e$ is a Lagrange multiplier.
Since the solution $\widehat{\Lm}$ must correspond to a connected graph, the effective resistances $\hat{r}_e >0$ for $e \in [p]^2$, thus in particular for every $e \in \Ec$, 
$\hat{r}_e = h_e - \hat{\lambda}_e > 0$.
Therefore $h_e > \hat{\lambda}_e \geq 0$ for every $e \in \Ec$,  which implies that $\Ec \subset \Ec_{\exists}$. Since $\Ec$ is connected, then $\Ec_{\exists}$ must also be connected.

\textbf{Part 2:}
We adapt the proof of Theorem 1 in \citet {ying2021minimax}. The main difference is that the edge set of  $\widehat{\Lm}$  is  contained in $\Ec\subset \Ec_{\exists}$. Hence, we need to prove that the objective function is lower bounded when $\Ec_{\exists}$ is connected, instead of assuming that $\Ec_{\exists}$ is the complete graph\footnote{\citet{ying2021minimax} showed that $h_e>0$ for all $e \in [p]\times [p]$ with probability one, when the data follows a L-GMRF. }.   We also use the matrix tree Theorem, which results in a much shorter proof.  
We start by  bounding the log determinant  term.
Let $\Ec, \Ec_{\exists}$ be connected graphs and $\Ec \subset \Ec_{\exists}$. Denote  the maximum weight by $\Vert \wv \Vert_{\infty} = \max_{e \in \Ec}w_e$, then we have  the inequalities
\begin{align*}
 &\sum_{\Tc \subset \Ec} \Omega(\Lm_{\Tc})   \\
 &\leq  \sum_{\Tc \subset \Ec} \Vert \wv \Vert_{\infty}^{p-1} = \Vert \wv \Vert_{\infty}^{p-1} \Omega(\Ec) \leq \Vert \wv \Vert_1^{p-1} \Omega(\Ec).
\end{align*}
Using the matrix tree Theorem,
\begin{equation*}
    \log\det\left( \Lm + \frac{1}{p}\Jm_p \right) \leq \log(p \Omega(\Ec)) + (p-1)\log(\Vert \wv \Vert_1).
\end{equation*}
To bound the linear term   we use  $\tr(\Lm \Sm) = \sum_{e \in \Ec} w_e h_e \geq  \underline{h} \Vert \wv \Vert_1$, where $\underline{h} = \min_{e \in \Ec} h_e$. 
Combining both inequalities, we have
\begin{equation}\label{eq_lower_bound_objective_w}
  \ell^s(\Lm,\Sm^{\dagger}) \geq  - \log(\Vert \wv \Vert_1) +\frac{\underline{h}}{p-1}\Vert \wv \Vert_1 + C,
\end{equation}
where $C$ is a constant that does not depend on $\wv$.  The hypothesis $\Ec \subset \Ec_{\exists}$ implies that $\underline{h}>0$.
Then,   the  expression in the right hand side of \eqref{eq_lower_bound_objective_w} is minimized when $\Vert \wv \Vert_1 = (p-1)/\underline{h}$, and since $\underline{h}>0$, 
\begin{equation}\label{eq_lower_bound_objective}
    \ell^s(\Lm,\Sm^{\dagger}) \geq \log(\underline{h})-\log(p-1) +1 + C   > - \infty.
\end{equation}
 \section{Consistency}
\label{sec:consis}
 We study  consistency in  the symmetrized  Stein loss:
\begin{align}\label{eq_sstein_definition}
    L^{ss}(\Lm_1, \Lm_2) &\coloneqq \frac{\ell^s(\Lm_1, \Lm_2) + \ell^s(\Lm_2,\Lm_1)}{2} \\
&= \frac{\tr\left( (\Lm_1  - \Lm_2)( \Lm_2^{\dagger} - \Lm_1^{\dagger}) \right)}{2(p-1)}. \label{eq_sstein_definition2}
\end{align}
Note that consistency in the symmetrized loss, immediately implies consistency in the non-symmetric loss, since $\ell^s(\Lm_1, \Lm_2) \leq 2 L^{ss}(\Lm_1, \Lm_2)$. The symmetrized Stein loss for positive definite matrices  gives equal treatment to estimation of the precision and covariance matrices \citep{stein1975estimation,ledoit2018optimal,soloff2020covariance}. For CGL matrices, the proposed symmetrized loss is invariant to pseudo inversion, since $ L^{ss}(\Lm_1, \Lm_2) =  L^{ss}(\Lm_1^{\dagger}, \Lm_2^{\dagger})$.
In addition,  the symmetrized Stein loss for CGL matrices  gives equal treatment to the edge weights and effective resistances, since we can alternatively write
\begin{equation}\label{eq_sstein_definition3}
  L^{ss}(\Lm_1, \Lm_2) =   \sum_{\mathclap{e \in \Ec_1 \cup \Ec_2}}  \frac{(w_{1,e} - w_{2,e})(r_{2,e} - r_{1,e})}{2(p-1)}. 
\end{equation}
where $\Ec_i$, $w_{i,e}$ and $r_{i,e}$ are the edge set, edge weights, and effective resistances of $\Lm_i$, respectively.

We make some assumptions over the data distribution, standard in the covariance estimation literature \citep{vershynin2018high,bunea2015sample,lounici2014high}.
\begin{assumption}\label{assump_subgauss}
Let $\Xm = [  \xv_1,\cdots \xv_n  ]$.
\begin{enumerate}
    \item The columns of $\Xm$ are zero mean  i.i.d. copies of the vector $\xv$ with covariance matrix $\Sigmam^{\star}=(\Lm^{\star})^{\dagger}$, where $\Lm^{\star}$ is the CGL of a connected graph $\Gc^{\star} = (\Vc, \Ec^{\star})$, with edge weights $w_e^{\star}$ for $e \in \Ec^{\star}$.
    \item $\xv$ is a sub-Gaussian vector, so there is constant $c_0$ so that for all $\uv$, then  $\E[ \vert \uv^{\top} \xv \vert^r]^{1/r} \leq c_0 \Vert  \uv^{\top} \xv\Vert_{\Psi_2}\sqrt{r}$
    \item There is a constant $c_1$ so that for all $\uv$, then $\Vert  \uv^{\top} \xv\Vert^2_{\Psi_2} \leq \E[(\uv^{\top} \xv)^2]/c_1$.
\end{enumerate}
\end{assumption}
Here $\Vert \cdot  \Vert_{\Psi_2}$ denotes the  sub-Gaussian norm \citep{vershynin2018high}.
%Our analysis involves the  following parameters:
%\begin{align}
%%R(\Ec) &\coloneqq \max_{e \in \Ec}r_e^{\star} = \max_{e \in \Ec}(\Sigma^{\star}_{ii} +\Sigma^{\star}_{jj} - 2\Sigma^{\star}_{ij}),\\
%r(\Ec) &\coloneqq \min_{e \in \Ec}r_e^{\star}=\min_{e \in \Ec}(\Sigma^{\star}_{ii} +\Sigma^{\star}_{jj} - 2\Sigma^{\star}_{ij}),\\
%\gamma(\Ec) &\coloneqq \frac{R(\Ec)}{r(\Ec)}.
%\end{align}
%
%
%
% A particular consequence of this is that 
% \begin{equation}
%     \Vert  \gv_e^{\top} \xv\Vert^2_{\Psi_2} \leq  r^{\star}_e/c_1.
% \end{equation}
Now we are in position to state our main result.
\begin{theorem}\label{th_consistency}
Let  $\Xm$   satisfy Assumption \ref{assump_subgauss}.    If  $\Ec^{\star} \subset \Ec \subset \Ec_{\exists}$, and    $n \geq c'\log(p)$, then 
\begin{equation}
 \Prob\left( L^{ss}(\widehat{\Lm}, \Lm^{\star})\leq C'   \sqrt{\frac{\log(p)}{n}} \right) \geq 1 - 2/p^2,
\end{equation}
for some universal  constants $c',C'$    that depend only on Assumption \ref{assump_subgauss}.
\end{theorem}
\autoref{th_consistency} provides consistency as long as $\log(p) = \Oc(n)$. In particular, we can consider the setting when $p,n \rightarrow \infty$ and $\frac{p}{n} \rightarrow \eta > 0$, where we have that
\begin{equation}\label{eq_high_dim_consistent}
 L^{ss}(\widehat{\Lm}, \Lm^{\star}) \rightarrow 0
\end{equation}
with  probability one.

Similar error rates have  appeared in other studies. \citet{rabbat2017inferring} proposed an estimator for the CGL that is robust to noise, and obtained a convergence rate in entry wise maximum norm in the order $\Oc(s\sqrt{log(p)/n})$,  where $s$ is the graph sparsity.  \citet{ying2020nonconvex} proposed a non convex CGL estimator that attains a rate $\Oc(\sqrt{s\log(p)/n})$ in Frobenius norm. More recently \cite{ying2021minimax} proposed an adaptive estimator attaining upper and lower bounds, also for the Frobenious norm, in the order of $\Oc(\sqrt{d\log(p)/n})$, where $d$ is the maximum unweighted degree. In these papers, the parameters $d$,  and $s$ depend on the graph structure, and may scale with the dimension. Note that while these bounds appear similar to our bound from \autoref{th_consistency} they are not comparable, since these papers studied consistency using matrix norms, while in this work we used the symmetrized Stein loss. 

Recently, \citet{soloff2020covariance} proved consistency of the GGLE in the symmetrized Stein loss, with rate $\Oc((1-\rho)^{-1}\sqrt{\log(p)/n})$, where $\rho = \max_{i \neq  j} \Sigma^{\star}_{ij}/\sqrt{\Sigma^{\star}_{ii} \Sigma^{\star}_{jj}}$.  For the GGLE, when $\rho \approx 1$ the error bound diverges. 
In contrast with previous works, when using the symmetrized Stein loss to measure consistency of  the  CGLE estimator  \eqref{eq_optimal_proposed}, we showed that the set $\Lc_p(\Ec)$ provides sufficient regularization to achieve consistency in high dimensions with  rate $\Oc(\sqrt{\log(p)/n})$. Moreover, this rate does not depend on the graph structure or population parameters such as $\rho$, $d$ or $s$, that may scale with the dimension. 
% \begin{remark}
% We can bound the symmetrized Stein loss under further assumptions, and some bounds for other norms. 
% \end{remark}

When the population CGL $\Lm^{\star}$ is a tree with known edge set $\Ec^{\star}$, the edge weights can be estimated with a   faster convergence rate. 
\begin{theorem}\label{th_consistency_trees}
Let  $\Xm$  satisfy Assumption \ref{assump_subgauss}.    If $\Ec^{\star}$ is a tree and $\Ec^{\star}=\Ec \subset \Ec_{\exists}$,  for  $n \geq c''\log(p)$, we have that 
\begin{equation}
 \Prob\left( L^{ss}(\widehat{\Lm}, \Lm^{\star})\leq C''   \frac{\log(p)}{n} \right) \geq 1 - 2/p^2,
\end{equation}
for some universal  constants $c'',C''$    depending only on Assumption \ref{assump_subgauss}.
\end{theorem}
The boosted  $\Oc(\log(p)/n)$ rate is possible because the population edge set is known, which allows us to exploit some results from \citet{pavez2019efficient,lu2018learning} for trees. It still remains an open question whether the rate  from \autoref{th_consistency} can be improved by using knowledge about the graph topology.  In the next subsections we present the proofs of \autoref{th_consistency} and \autoref{th_consistency_trees}, and discuss potential paths for improvements. 
% \textcolor{red}{Discussions regarding the meta-review, and other reviewers comments. 
% \begin{enumerate}
%     \item Optimality of bound, and dpendence on the edge set. State that
% \end{enumerate}
% }
\begin{figure*}[htb]
%\vspace{.3in}
\begin{center}
  \begin{subfigure}[b]{0.24\textwidth}
  \centering
\includegraphics[width=0.6\textwidth]{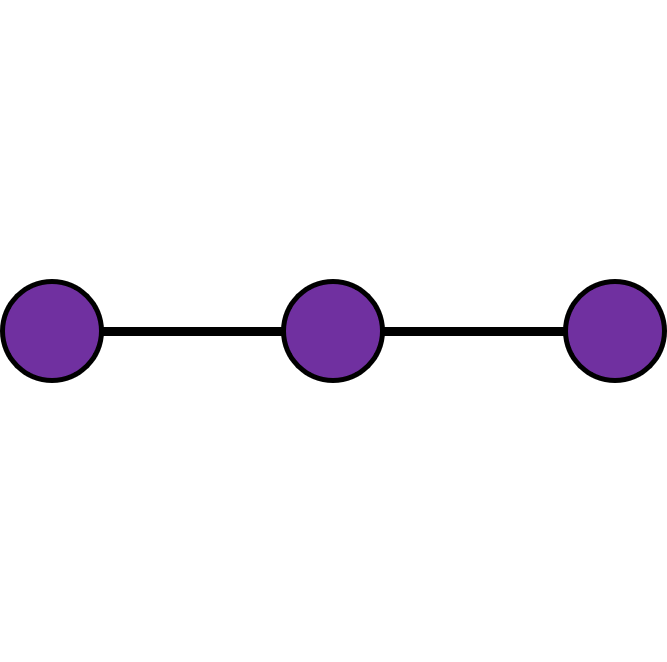}
\caption{}
\label{fig:graphs:path}
\end{subfigure}
\begin{subfigure}[b]{0.24\textwidth}
\centering
\includegraphics[width=0.6\textwidth]{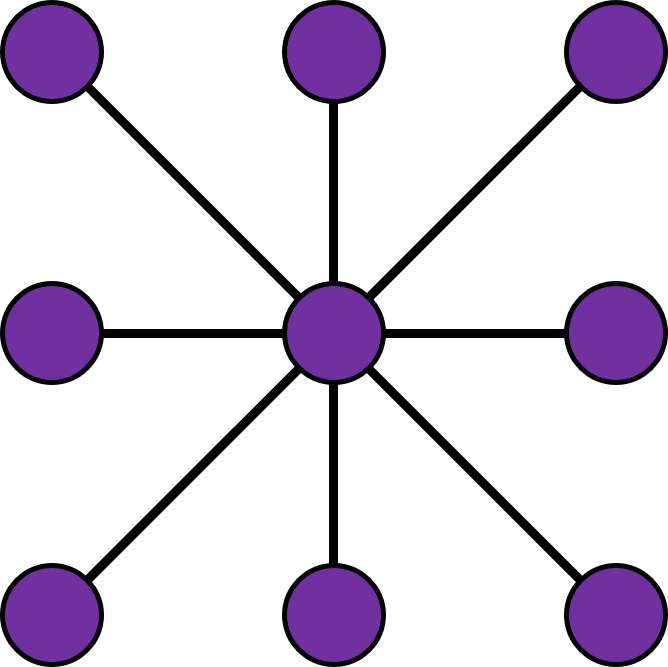}
\caption{}
\label{fig:graphs:star}
\end{subfigure}
\begin{subfigure}[b]{0.24\textwidth}
\centering
\includegraphics[width=0.6\textwidth]{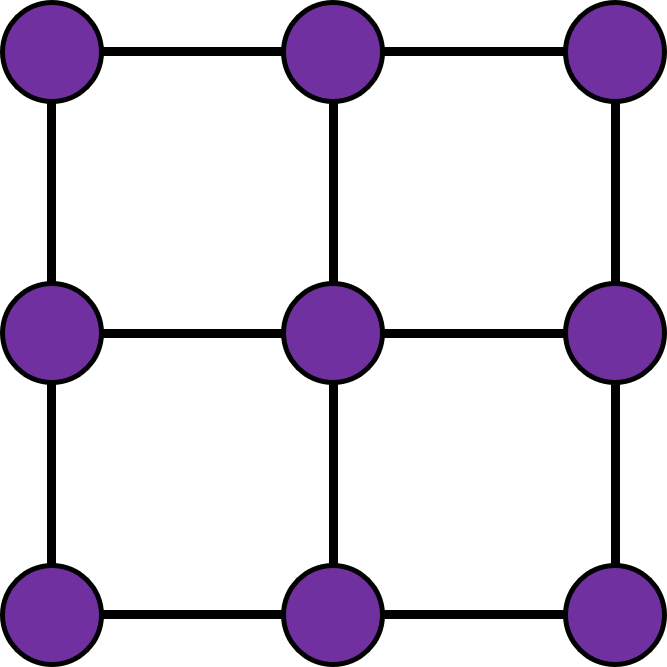}
\caption{}
\label{fig:graphs:4conn}
\end{subfigure}
%\begin{subfigure}[b]{0.19\textwidth}
%\centering
%\includegraphics[width=0.7\textwidth]{figs/8conn.png}
%\caption{}
%\label{fig:graphs:8conn}
%\end{subfigure}
\begin{subfigure}[b]{0.24\textwidth}
\centering
\includegraphics[width=0.6\textwidth]{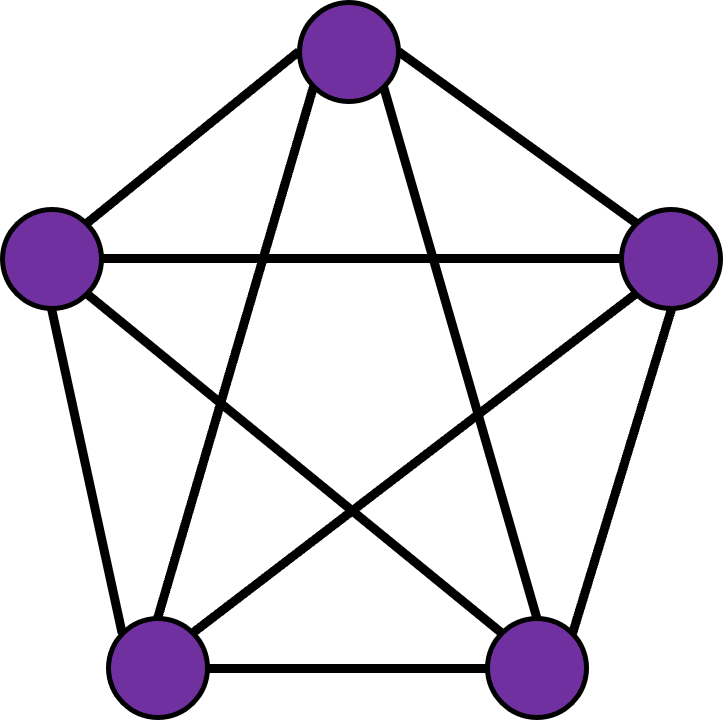}
\caption{}
\label{fig:graphs:comp}
\end{subfigure}  
\end{center}
%\vspace{.3in}
\caption{Graph types used in experiments. Figures \ref{fig:graphs:path} and \ref{fig:graphs:star} correspond to trees,  a path  with $p=3$ nodes, and a star  with $p=9$ nodes, respectively. Figure \ref{fig:graphs:4conn}  depicts a regular grid graph with maximum degree $4$, and $p=9$ nodes. Figure \ref{fig:graphs:comp} is the complete graph with $p=5$ nodes.  }
\label{fig:graphs}
\end{figure*}
\begin{figure*}[htb]
\centering
\vspace{.3in}
\begin{subfigure}[b]{0.25\textwidth}
\includegraphics[width=1\textwidth]{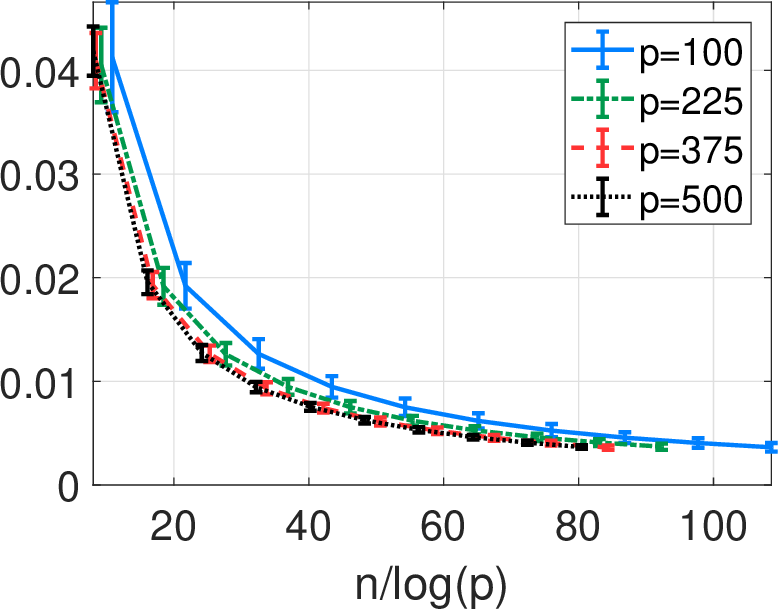}
\caption{Path}
\end{subfigure}% %this comment is super important for side by side 
\begin{subfigure}[b]{0.25\textwidth}
\includegraphics[width=1\textwidth]{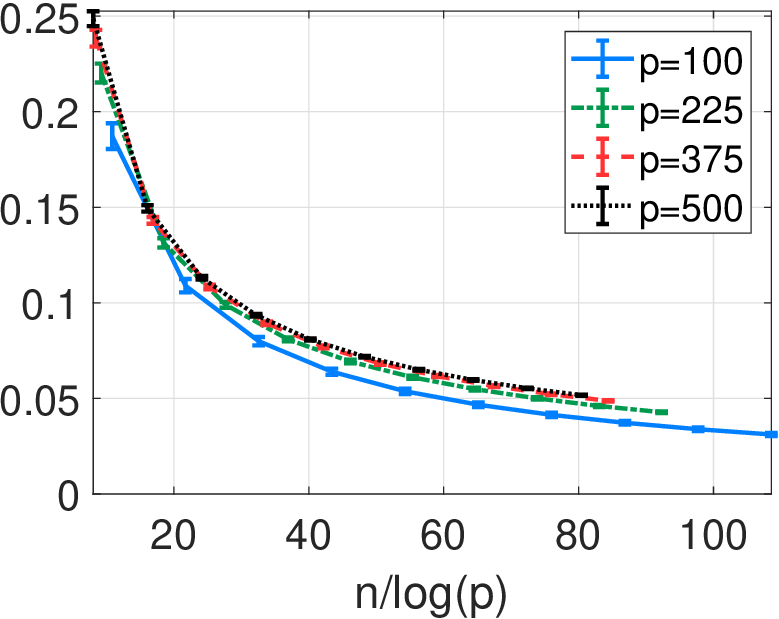}
\caption{Star}
\end{subfigure}%
\begin{subfigure}[b]{0.25\textwidth}
\includegraphics[width=1\textwidth]{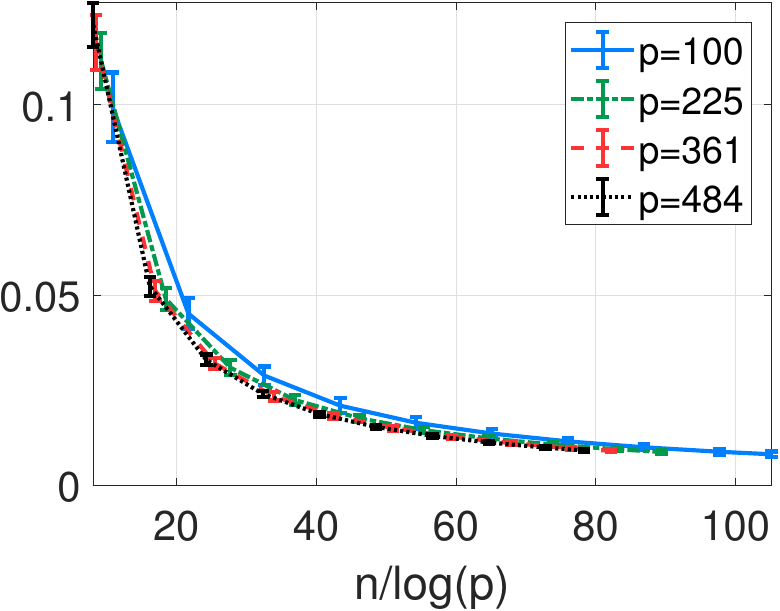}
\caption{Grid}
\end{subfigure}%
\begin{subfigure}[b]{0.25\textwidth}
\includegraphics[width=1\textwidth]{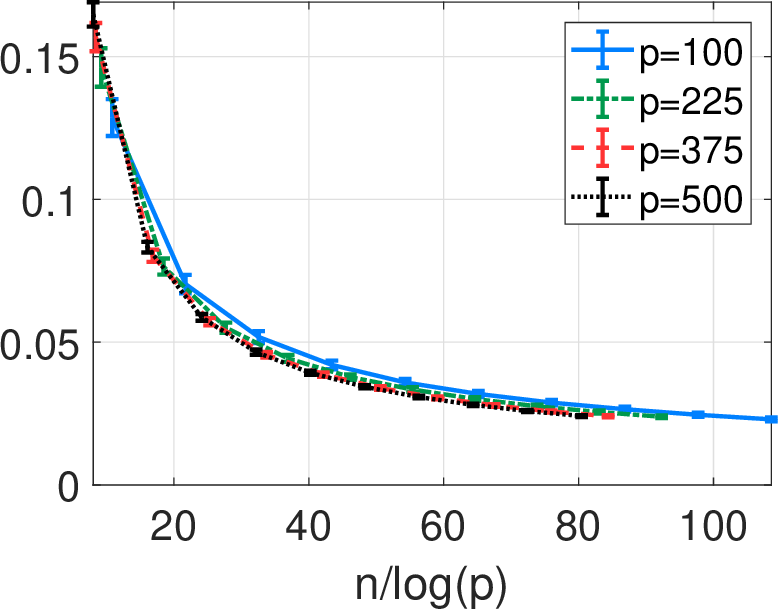}
\caption{Complete}
\end{subfigure}
%\vspace{.3in}
\caption{Convergence of the symmetrized Stein loss for various graph types as a function of $n/log(p)$.}
\label{fig:convergence1}
\end{figure*}
\begin{figure*}[htb]
\centering
\vspace{.3in}
\begin{subfigure}[b]{0.25\textwidth}
\includegraphics[width=1\textwidth]{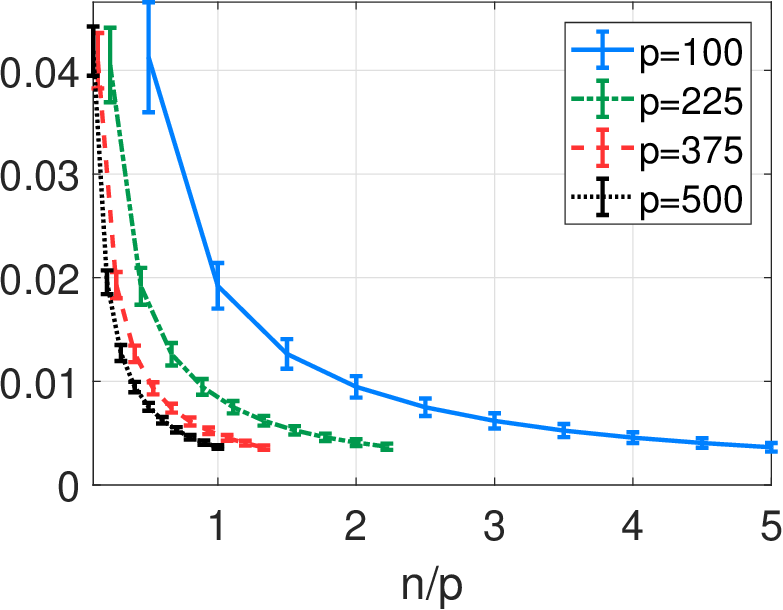}
\caption{Path}
\end{subfigure}% %this comment is super important for side by side 
\begin{subfigure}[b]{0.25\textwidth}
\includegraphics[width=1\textwidth]{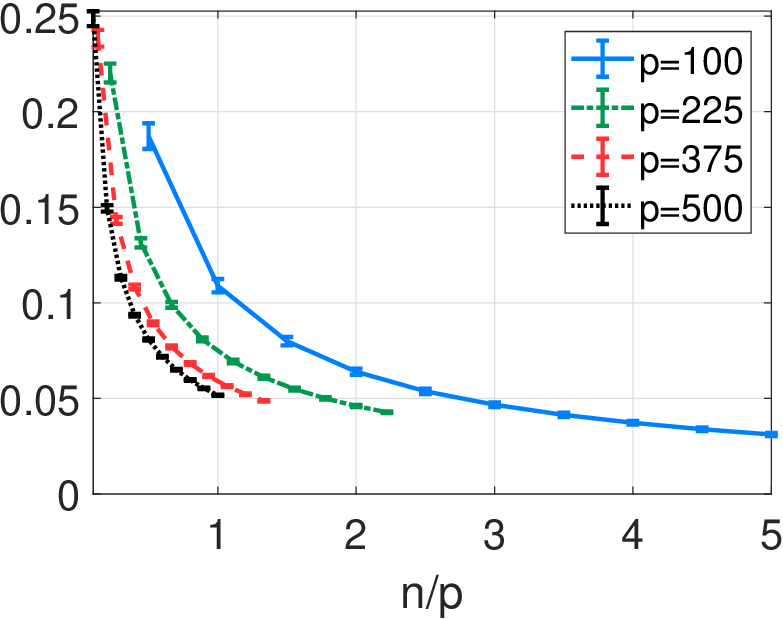}
\caption{Star}
\end{subfigure}%
\begin{subfigure}[b]{0.25\textwidth}
\includegraphics[width=1\textwidth]{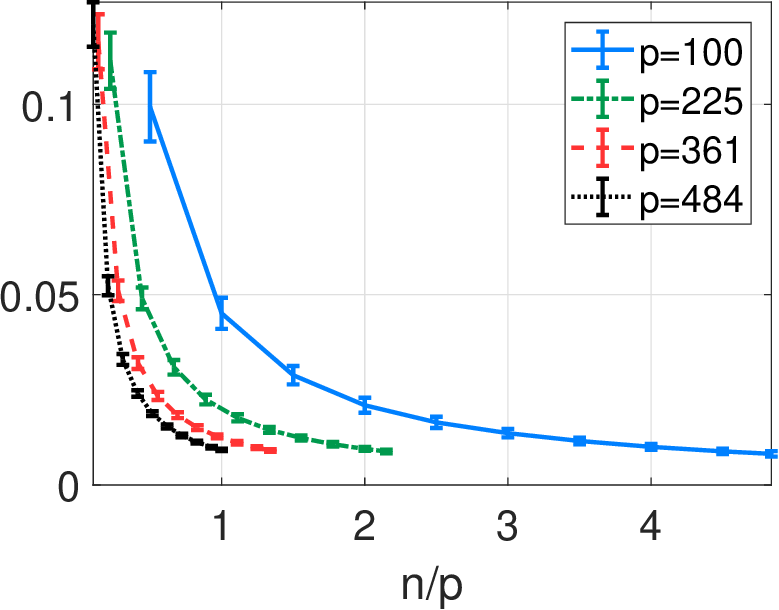}
\caption{Grid}
\end{subfigure}%
\begin{subfigure}[b]{0.25\textwidth}
\includegraphics[width=1\textwidth]{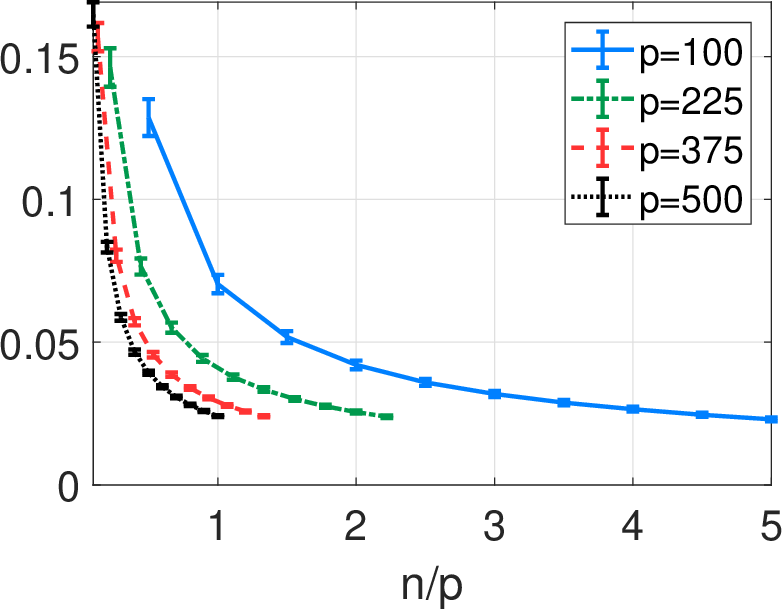}
\caption{Complete}
\end{subfigure}
%\vspace{.3in}
\caption{Convergence of the symmetrized Stein loss for various graph types as a function of $n/p$.}
\label{fig:convergence2}
\end{figure*}
%%%%%%%%%%%%%%%%%%%%%%%%%%%%%%%
\begin{figure*}[htb]
\centering
\vspace{.3in}
\begin{subfigure}[b]{0.3\textwidth}
\includegraphics[width=1\textwidth]{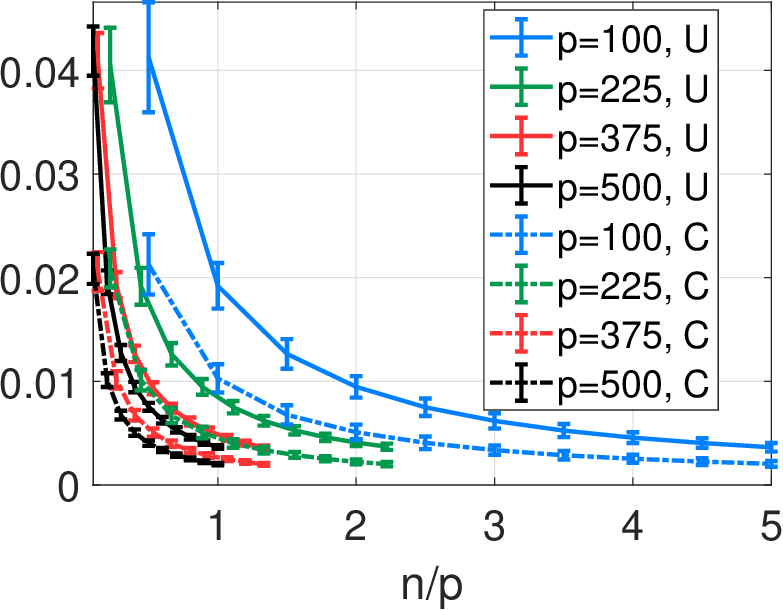}
\caption{Path}
\end{subfigure}% %this comment is super important for side by side 
\begin{subfigure}[b]{0.3\textwidth}
\includegraphics[width=1\textwidth]{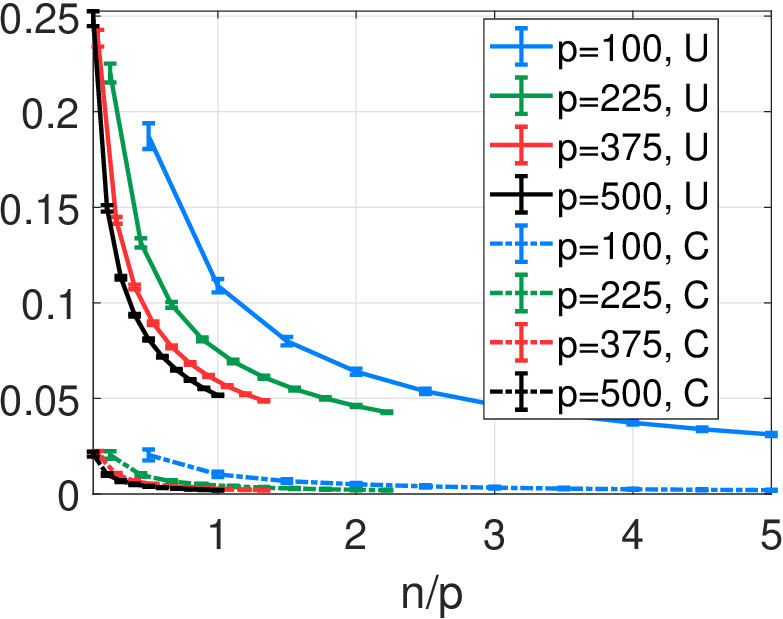}
\caption{Star}
\end{subfigure}%
\begin{subfigure}[b]{0.3\textwidth}
\includegraphics[width=1\textwidth]{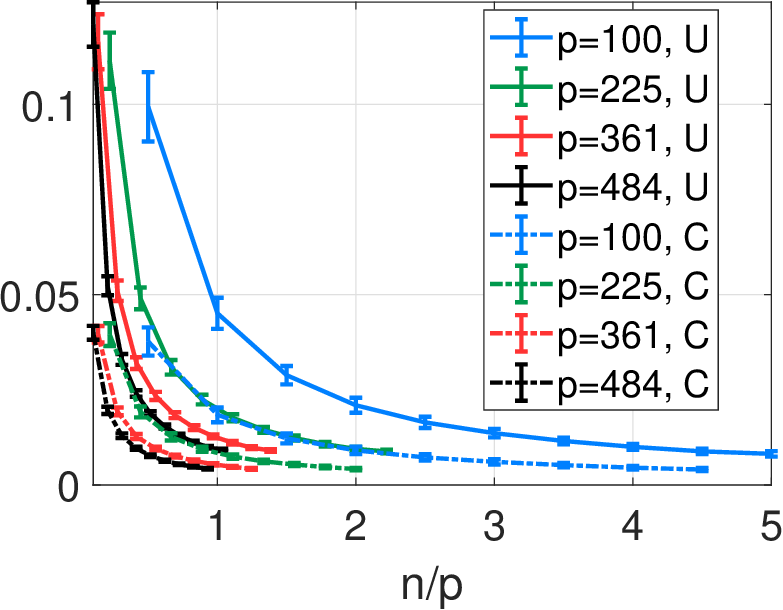}
\caption{Grid}
\end{subfigure}
%\vspace{.3in}
\caption{Comparison of edge constrained (C) and unconstrained (U) CGLE. Plots show convergence of the symmetrized Stein loss for various graph types as a function of $n/p$.}
\label{fig:convergence3}
\end{figure*}
\subsection{Proof of \autoref{th_consistency}}
We first derive a  fundamental inequality for $L^{ss}$. Let 
\begin{equation}\label{eq_delta_E}
\Delta(\Ec) \coloneqq \max_{e \in \Ec} \left\vert 1  - \frac{h_e}{r_e^{\star}}  \right\vert,
\end{equation}
where $h_e$ is defined in \eqref{eq_he}, and $r^{\star}_e$ is the effective resistance of the population Laplacian $\Lm^{\star}$. The first part of the proof establishes the following inequality for $L^{ss}$.
\begin{proposition}\label{prop_master_inequality}
The symmetrized Stein loss satisfies
\begin{equation}
L^{ss}(\widehat{\Lm},\Lm^{\star}) \leq \Delta(\Ec)(3/2 + L^{ss}(\widehat{\Lm},\Lm^{\star})).
\end{equation}
\end{proposition}
Proposition \ref{prop_master_inequality} is the key element of the proof, since  conditioned on the event $\Delta(\Ec) < 1/2$,  it  implies
\begin{equation}\label{eq_master_inequality2}
L^{ss}(\widehat{\Lm},\Lm^{\star})  \leq 3\Delta(\Ec).
\end{equation}
The fact that the quantity $\Delta(\Ec)$ is normalized,  results in the graph independent bound from \autoref{th_consistency}.

For the second part of the proof, we bound  $\Delta(\Ec)$  using  a concentration inequality by \citet{jankova2015confidence}. The required result is presented in  Lemma \ref{lemma_concentration_eff_resistance2} which can be found in \autoref{ssec_aux_lemmas}. 
From Lemma \ref{lemma_concentration_eff_resistance2}, we have that   $\Delta(\Ec) \leq 4C\sqrt{2\log(p)/n} <1/2 $ with probability at least $1-1/p^2$,  whenever
\begin{equation}
n \geq (64 C^2 \vee 1) 2\log(p),
\end{equation}
where $C$ is a constant that depends on Assumption \ref{assump_subgauss}.
Combining this last display with \eqref{eq_master_inequality2} results in 
\begin{equation}
L^{ss}(\widehat{\Lm},\Lm^{\star})  \leq 3\Delta(\Ec) \leq 12C  \sqrt{\frac{2\log(p)}{n}}.
\end{equation}
\begin{remark}
Proposition \ref{prop_master_inequality} resembles a bound obtained by \cite{soloff2020covariance} for the symmetrized Stein loss between GGL matrices. In their case, however there is an additional constant $(1-\rho)^{-1}$ (see \autoref{tab_existence}) that depends on the population covariance matrix, and diverges as $\rho \rightarrow 1$. In our case we are able to avoid such type of constants by exploiting properties of CGL matrices that are not shared by GGL matrices, namely the identity \eqref{eq_sstein_definition3}.
\end{remark}
Now we prove Proposition \ref{prop_master_inequality}. 
%\subsubsection{Proof of Proposition \ref{prop_master_inequality}}
Using \eqref{eq_sstein_definition3},   we have that
\begin{equation}
	L^{ss}(\widehat{\Lm},\Lm^{\star}) 
	= \frac{1}{2(p-1)} \sum_{e \in \Ec}(\hat{w}_e - w_e^{\star})(r_e^{\star} - \hat{r}_e).\label{eq_sstein_weight_resist_formula}
\end{equation}
We bound the sum in \eqref{eq_sstein_weight_resist_formula} using the KKT conditions,
\begin{align} \nonumber 
& \sum_{e \in \Ec} (\hat{w}_e - w_e^{\star})( r_e^{\star} - \hat{r}_e)   \\ \nonumber 
&=\sum_{e \in \Ec} (\hat{w}_e - w_e^{\star})(r_e^{\star}  - h_e) + \sum_{e \in \Ec} (\hat{w}_e - w_e^{\star})(h_e - \hat{r}_e)\\ \nonumber 
&=\sum_{e \in \Ec} (\hat{w}_e - w_e^{\star})(r_e^{\star}  - h_e) - \sum_{e \in \Ec}  w_e^{\star}(h_e - \hat{r}_e) \\ 
&\leq \sum_{e \in \Ec} (\hat{w}_e - w_e^{\star})(r_e^{\star}  - h_e). \label{eq_sstein_inequality1}
\end{align}
In the second equality we combine  \eqref{eq_kkt_1} and \eqref{eq_kkt_2},  and use $\hat{w}_e(h_e - \hat{r}_e) = 0$. The  inequality \eqref{eq_sstein_inequality1} is obtained after combining \eqref{eq_kkt_1} and \eqref{eq_kkt_3} producing $(h_e - \hat{r}_e) \geq 0$. 
For connected graphs, all effective resistances are positive, thus  each term in the sum can be normalized, and  \eqref{eq_sstein_inequality1}  can be rewritten as
\begin{equation}
\sum_{e \in \Ec} (\hat{w}_e r_e^{\star} - w_e^{\star} r_e^{\star})\left(1  - \frac{h_e}{r_e^{\star}}\right).
\end{equation}
Then we have the following sequence of inequalities
\begin{align}
\eqref{eq_sstein_inequality1} 
&\leq \max_{e \in \Ec} \left\vert 1  - \frac{h_e}{r_e^{\star}}  \right\vert \sum_{e \in \Ec} \vert\hat{w}_e r_e^{\star} - w_e^{\star} r_e^{\star}\vert \label{eq_sstein_potential_improvement1}\\
&\leq \Delta(\Ec)  \sum_{e \in \Ec} (\hat{w}_e r_e^{\star} + w_e^{\star} r_e^{\star})\label{eq_sstein_potential_improvement2}
\end{align}
The second inequality  above uses \eqref{eq_delta_E}, and the triangular inequality. Since for any connected graph $p-1=\tr(\Lm \Lm^{\dagger}) = \sum_{e \in \Ec}w_e r_e$, and $\tr(\widehat{\Lm}\Sigmam^{\star})=\sum_{e\in \Ec}\hat{w}_e r^{\star}_e$, we have that $\sum_{e \in \Ec} (\hat{w}_e r_e^{\star} + w_e^{\star} r_e^{\star})
    =\tr(\widehat{\Lm}\Sigmam^{\star}) + p-1$, thus
\begin{equation}
2(p-1)L^{ss}(\widehat{\Lm}, \Lm^{\star})\leq \Delta(\Ec)(\tr(\widehat{\Lm}\Sigmam^{\star}) + p-1)
\end{equation}
We conclude by applying
   $\tr(\widehat{\Lm}\Sigmam^{\star}) \leq 2(p-1)(1+L^{ss}(\widehat{\Lm},\Lm^{\star}))$, which is a direct consequence of \eqref{eq_sstein_definition2}, and the fact that $\tr(\Lm^{\star} (\widehat{\Lm})^{\dagger}) \geq 0$.
\begin{remark}
In the proof of Proposition \ref{prop_master_inequality}, when we go from  \eqref{eq_sstein_potential_improvement1} to \eqref{eq_sstein_potential_improvement2} with the triangular inequality, a the  term $\sum_{e \in \Ec} \vert\hat{w}_e r_e^{\star} - w_e^{\star} r_e^{\star}\vert$ that vanishes to zero as $n \rightarrow \infty$,  is bounded by  another term $\sum_{e \in \Ec} (\hat{w}_e r_e^{\star} + w_e^{\star} r_e^{\star})$ that converges to a positive constant. Therefore, if the bound from \eqref{eq_sstein_potential_improvement2} can be improved, then the gap between \autoref{th_consistency} and \autoref{th_consistency_trees} may be reduced.
\end{remark}

%%%%%%
\subsection{Proof of \autoref{th_consistency_trees}}
For trees, the effective resistances on the edge set are the inverse of the edge weighs, thus  $r_e = 1/w_e$ for $e \in  \Ec$. The CGLE for a tree with edge set $\Ec$ also has a closed form solution \citep{lu2018learning} given by $\hat{w}_e = 1/h_e$ for each $e \in \Ec$, which implies that $\hat{r}_e = h_e$ for $e \in \Ec$. Using these identities,   the symmetrized Stein loss has an alternative expression given by:
\begin{equation}
    L^{ss}(\widehat{\Lm}, \Lm^{\star}) = \frac{1}{2(p-1)}  \sum_{e\in \Ec} \hat{w}_e r_e^{\star} \left( \frac{\hat{r}_e -  r_e^{\star}}{  r_e^{\star}} \right)^2.
\end{equation}
Following a similar procedure as in the proof of Proposition \ref{prop_master_inequality}, we can derive the inequality 
\begin{align}
  L^{ss}(\widehat{\Lm}, \Lm^{\star}) \leq \Delta^2(\Ec)(1 +  L^{ss}(\widehat{\Lm}, \Lm^{\star})). 
\end{align}
Conditioned on the event $\Delta^2(\Ec)<1/2$, the above inequality implies 
\begin{equation}
 L^{ss}(\widehat{\Lm}, \Lm^{\star}) \leq 2\Delta^2(\Ec).   
\end{equation}
The proof is concluded by applying a concentration inequality to $\Delta(\Ec)$, similar to the proof of \autoref{th_consistency}. Details are provided in \autoref{ap_proof_trees}.
\section{Experiments}
\label{sec:exp}
In this section we perform numerical simulations to validate our theoretical results. 
We consider $4$ different types of graph that can be easily generated with an arbitrary number of nodes $p$.  In \autoref{fig:graphs:path} and \autoref{fig:graphs:star} we depict two types of trees,  a path graph  with $p=3$ nodes, and a star graph with $p=9$ nodes, respectively. We will also consider a $\sqrt{p} \times \sqrt{p}$ grid graph (Figure \ref{fig:graphs:4conn}), with maximum degree equal to $4$,  and the complete graph (\autoref{fig:graphs:comp}). Since the path and star graphs are trees, they  have $p-1$ edges. The $\sqrt{p} \times \sqrt{p}$ grid graph has  $2\sqrt{p}(\sqrt{p}-1)$   edges. The complete graph has $p(p-1)/2$ edges. 
For the path, star and complete graphs we set $p \in \lbrace 100, 225, 375, 500 \rbrace$, and for the grid graph we set $p \in \lbrace 100, 225, 361, 484  \rbrace$, since we require $\sqrt{p}$ to be integer. All graphs have  edge weights $w^{\star}_e=1$. 

For each graph type and dimension $p$, we construct the CGL matrix $\Lm^{\star}$, and generate $T=100$ data matrices $\Xm$ of size $p \times 500$, with i.i.d. Gaussian columns, zero mean and covariance matrix $(\Lm^{\star})^{\dagger}$.

\subsection{Estimation without  topology constraints}
First we evaluate the performance of the CGLE when the constraint edge set is the complete graph, hence, there is no prior information about the true graph topology.  We take several values of  of $n \in [50, 500]$, compute the sample covariance matrix and calculate the CGLE using the coordinate descent algorithm proposed by \cite{pavez2019efficient}({\url{https://github.com/STAC-USC/graph_learning_CombLap}}), using their default parameters. For each graph, and pair $(p,n)$ we repeat the experiment $T$ times. In \autoref{fig:convergence1} we show the average values of $L^{ss}$,  with error bars of $1$ standard deviation,  as a function of $n/\log(p)$. Note that for each graph type, all curves clump together, in agreement with the weak dependence on the dimension predicted by \autoref{th_consistency}. In 
\autoref{fig:convergence2}, we  show the same data as in \autoref{fig:convergence1}, but now as a function of $n/p$. We observe that for all  graph types, and  for each value of $n/p$, the symmetrized Stein loss approaches zero faster in higher dimensions, as predicted in \eqref{eq_high_dim_consistent}. 
%

%%%%%%%%%%%%%%%%%%%%%%%%%%%%%%%%%%%%

\subsection{Estimation with topology constraints}
We compare the CGLE with and without knowledge of the true graph topology $\Ec^{\star}$. We only consider  the sparse graphs (path, star and grid).  \autoref{fig:convergence3} displays the symmetrized Stein loss as a function of $n/p$. 
For all graphs, and all pairs $(n,p)$,  graph structure constraints reduces the loss. 
In the unconstrained case (solid lines in \autoref{fig:convergence3}), the star graph has the highest loss, and takes longer to converge, while the path graph  is the fastest. Once the tree  constraint is added, the path and star graphs   have  similar fast convergence behavior. 
This similitude may be attributed to the values taken by the effective resistances of edges and non edges. For the path, $r^{\star}_e=1$ for edges, while is takes values between $2$ and $p-1$ for non edges. For the grid graph, the edge effective resistances are smaller than $1$, while the non edge resistances can be much larger. For the star graph, the edge resistances are equal to $1$, while all the non edge resistances are equal to $2$. When $n$ is smaller, it is more likely that a edge and non edge resistances are similar for the star graph, than for the path and grid graphs, resulting    more difficult to obtain an accurate estimation of the CGL, and distinguish between edges and non edges.
\section{Conclusion}
\label{sec:conc}
In this paper we proved that the combinatorial graph Laplacian estimator with edge constraints exists, if and only if, a certain data dependent graph is connected. We proposed the  symmetrized Stein loss for CGL matrices of connected graphs. We showed that the CGLE is consistent under this loss, even in high dimensions. Unlike previous consistency results for various graph Laplacian estimators, our proof  does not require (non) convex   regularization, sparsity, or other type of structure. For tree structured graphs, when the graph topology is known, we can achieve consistency at a faster rate. 
Interesting directions for future work include improvements to the error  bounds   that exploit the graph structure to narrow the gap between the convergence rates from  \autoref{th_consistency} and \autoref{th_consistency_trees}, analysis of other estimators for the CGL using the symmetrized Stein loss, and determining  optimal convergence rates and matching lower bounds. 
\subsubsection*{Acknowledgements}
The author thanks Antonio Ortega for useful discussions. 
This work was funded by NSF under grants CCF-1410009 and CCF-2009032.
%All acknowledgments go at the end of the paper, including thanks to reviewers who gave useful comments, to colleagues who contributed to the ideas, and to funding agencies and corporate sponsors that provided financial support. 
%To preserve the anonymity, please include acknowledgments \emph{only} in the camera-ready papers.

%\subsubsection*{References}
%\vfill
\bibliography{refs}

%%%%%%%%%%%%%%%%%%%%%%%%%%%%%%%%%%%
%%%%%% SUPPLEMENT (OPTIONAL) %%%%%%
%%%%%%%%%%%%%%%%%%%%%%%%%%%%%%%%%%%

\clearpage
\appendix

\thispagestyle{empty}

% For one-column format, uncomment the following:
\onecolumn \makesupplementtitle
% For two-column format, uncomment the following:
%\twocolumn[ \makesupplementtitle ]

\section{Bounds on edge weights of GGLE}
\label{ap_bound_GGL}
Consider the GGL  estimation problem, with $\alpha \geq 0$  
\begin{equation}\label{eq_GGL_estimation}
    \min_{\Theta_{ij} \leq 0, i \neq j}-\log\det(\Thetam) + \tr(\Thetam \Sm) + \alpha \sum_{i \neq j} \vert \Theta_{ij} \vert.
\end{equation}
Let $\Km = \Sm + \alpha(\Id - \mathbf{1}\mathbf{1}^{\top})$, and $\Lambdam = (\lambda_{ij})$ with $\lambda_{ii}=0$ for all $i$. The KKT optimality conditions are 
\begin{align}
-\Thetam^{-1} + \mathbf{K}  + \mathbf{\Lambda} &=0 \label{kkteq1}\\
\lambda_{ij}\Theta_{ij}&=0,~\forall i \neq j \label{kkteq4}\\
\Theta_{ij} &\leq 0, ~\forall i \neq j\label{kkteq5}\\
\lambda_{ij} &\geq 0, ~\forall i \neq j\label{kkteq6} \\
\Thetam &\succ 0.
\end{align} 
Consider the set  $\Sc = \lbrace i,j \rbrace$. We can rewrite \eqref{kkteq1} for the sub-matrices indexed by $\Sc$, thus
\begin{equation}
    (\Thetam^{-1})_{\Sc,\Sc} = \Km_{\Sc,\Sc} + \Lambdam_{\Sc,\Sc}.
\end{equation}
Using the Schur complement we have that
\begin{equation}
    \Thetam_{\Sc,\Sc} - \Thetam_{\Sc,\Sc^c} (\Thetam_{\Sc^c,\Sc^c})^{-1} \Thetam_{\Sc^c,\Sc} = (\Km_{\Sc,\Sc} + \Lambdam_{\Sc,\Sc})^{-1}.
\end{equation}
Since $\Thetam_{\Sc^c,\Sc^c}$ is a GGL matrix, its inverse is non negative, thus the matrix  $\Thetam_{\Sc,\Sc^c} (\Thetam_{\Sc^c,\Sc^c})^{-1} \Thetam_{\Sc^c,\Sc}$ is also non negative (entrywise), and
\begin{equation}
 \Thetam_{\Sc,\Sc}  \geq  (\Km_{\Sc,\Sc} + \Lambdam_{\Sc,\Sc})^{-1}.
\end{equation}
Since $\Km_{\Sc,\Sc} + \Lambdam_{\Sc,\Sc}$ is a $2 \times 2$  matrix, we can compute its inverse in closed form, leading to the inequality
\begin{equation}
   \Theta_{ij} \geq \frac{-(K_{ij} + \lambda_{ij})}{K_{ii}K_{jj} - (K_{ij} + \lambda_{ij})^2}.
\end{equation}
When $\Theta_{ij}<0$, we have that $\lambda_{ij}=0$, and $K_{ij} \geq 0$, thus
\begin{equation}
    \vert \Theta_{ij} \vert \leq \frac{K_{ij} }{K_{ii}K_{jj} - K_{ij}^2}.
\end{equation}
The inequality of \autoref{tab_existence} follows after taking $\alpha=0$ and normalizing by $S_{ii}S_{jj}$.
\section{Auxiliary lemmas}
\label{ssec_aux_lemmas}
\begin{lemma}\label{lemma_concentration_eff_resistance}
Let each column of $\Xm$ be independent identically distributed, and satisfying Assumption \ref{assump_subgauss}, then for all $t>0$,
\begin{equation}
 \Prob(\Delta(\Ec) > C  (t + \sqrt{2t}) ) \leq 2\vert \Ec \vert \exp(-nt),
\end{equation}
For some universal constant $C$ depending on Assumption \ref{assump_subgauss}.
\begin{proof}
We write $\gv_e^{\top}(\Sigmam^{\star} - \Sm)\gv_e = r_e^{\star} - h_e$, and apply  Lemma 6 from \cite{jankova2015confidence},  then  for all $t>0$
\begin{equation}
    \Prob(\vert \gv_e^{\top}(\Sm - \Sigmam^{\star})\gv_e \vert > C r_e^{\star} (t + \sqrt{2t}) ) \leq 2\exp(-nt).
\end{equation}
  We conclude by taking an union bound.
\end{proof}
\end{lemma}

\begin{lemma}\label{lemma_concentration_eff_resistance2}
Under the assumptions of Lemma \ref{lemma_concentration_eff_resistance}, fix $\epsilon >0$, if
\begin{equation}
    n \geq (16C^2\epsilon^{-2} \vee 1) 2\log(p)
\end{equation}
then $\Delta(\Ec) \leq \epsilon$ with probability at least $1-2p^{-2}$.
\begin{proof}
Set $t = 4\log(p)/n$, and since $\vert \Ec \vert \leq p^2$, from  Lemma \ref{lemma_concentration_eff_resistance} we have that
\begin{equation}\label{eq_error_bound_lemma}
    \Delta(\Ec) \leq 2C \left( \frac{2\log(p)}{n}+ \sqrt{\frac{2\log(p)}{n}} \right),
\end{equation}
with probability at least $1 - 2p^{-2}$. We need to pick $n$, so that the RHS of \eqref{eq_error_bound_lemma} is below $\epsilon$. First note that since  $n \geq 2\log(p)$, then the RHS of \eqref{eq_error_bound_lemma} obeys
\begin{equation}
 \Delta(\Ec) \leq 4C  \sqrt{\frac{2\log(p)}{n}}.
\end{equation}
The desired claim follows since $n \geq 32 C^2\epsilon^{-2}  \log(p)$.
\end{proof}
\end{lemma}

\section{Proof of Theorem \ref{th_consistency_trees}}
\label{ap_proof_trees}
\begin{align}
  L^{ss}(\widehat{\Lm}, \Lm^{\star}) &\leq  \frac{\Delta(\Ec)^2}{2(p-1)} \sum_{e \in \Ec}\hat{w}_e r_e^{\star}\leq \Delta(\Ec)^2(1 +  L^{ss}(\widehat{\Lm}, \Lm^{\star})). \label{eq_master_inequality_trees}
\end{align}
The last display uses the relation $\sum_{e \in \Ec}\hat{w}_e r_e^{\star} = \tr(\widehat{\Lm}\Sigmam^{\star}) \leq 2(p-1)(1 +  L^{ss}(\widehat{\Lm}, \Lm^{\star}))$,  as in the proof of  \autoref{th_consistency}.
We can use Lemma \ref{lemma_concentration_eff_resistance}, and since the graph is a tree, $\Ec \leq p$, thus 
\begin{equation}
 \Prob(\Delta(\Ec) > C  (t + \sqrt{2t}) ) \leq 2 p \exp(-nt).
\end{equation}
If we pick $t = 3\log(p)/n$, we have that with probability at least $1-2/p^2$, 
\begin{equation}
\Delta(\Ec) \leq C\left( \frac{3\log(p)}{n} + \sqrt{\frac{6\log(p)}{n}} \right).
\end{equation}
Now for any $\epsilon>0$, if $n \geq 3(8C^2\epsilon^{-1} \vee 1)\log(p)$, then with probability greater than $1-2/p^2$
\begin{equation}
\Delta(\Ec) \leq 2C \sqrt{\frac{6\log(p)}{n}}\leq \epsilon.
\end{equation}
Now picking $\epsilon^2=1/2$, from \eqref{eq_master_inequality_trees} we obtain
\begin{equation}
 L^{ss}(\widehat{\Lm}, \Lm^{\star}) \leq 2 \Delta(\Ec)^2 \leq 48C^2\frac{\log(p)}{n}.
\end{equation}

\end{document}